\documentclass{article}

\usepackage{microtype}
\usepackage{graphicx}
\usepackage{booktabs}
\usepackage{subcaption}
\usepackage{wrapfig}
\usepackage{amsthm}
\usepackage{amsmath}
\usepackage{amsfonts}
\usepackage{amssymb}
\usepackage{color, colortbl}
\usepackage{float}
\usepackage[inline]{enumitem}
\usepackage{subfiles}
\usepackage[noend]{algpseudocode}
\usepackage{tikz}
\usepackage[all]{nowidow}
\usetikzlibrary{calc, shapes, positioning}

\usepackage{hyperref}
\usepackage{cleveref}

\usepackage[accepted]{icml2018}

\theoremstyle{definition}

\DeclareMathOperator*{\argmax}{arg\,max}
\definecolor{Gray}{gray}{0.9}
\definecolor{LightGray}{gray}{0.97}
\setlist[itemize]{leftmargin=*}
\newcommand*\Let[2]{\State #1 $\gets$ #2}
\algrenewcommand\alglinenumber[1]{{\sf\footnotesize#1}}
\algrenewcommand\algorithmicrequire{\textbf{Precondition:}}
\algrenewcommand\algorithmicensure{\textbf{Postcondition:}}
\algnewcommand{\LeftComment}[1]{\Statex \(\triangleright\) #1}

\icmltitlerunning{Learning Independent Causal Mechanisms}

\begin{document}

\twocolumn[
\icmltitle{Learning Independent Causal Mechanisms}



\icmlsetsymbol{equal}{*}

\begin{icmlauthorlist}
\icmlauthor{Giambattista Parascandolo}{tue,eth}
\icmlauthor{Niki Kilbertus}{tue,cam}
\icmlauthor{Mateo Rojas-Carulla}{tue,cam}
\icmlauthor{Bernhard Sch\"olkopf}{tue}
\end{icmlauthorlist}

\icmlaffiliation{tue}{Max Planck Institute for Intelligent Systems}
\icmlaffiliation{eth}{Max Planck ETH Center for Learning Systems}
\icmlaffiliation{cam}{University of Cambridge}

\icmlcorrespondingauthor{Giambattista Parascandolo}{gparascandolo@tue.mpg.de}

\icmlkeywords{learning, mechanisms, causal, causality, independent, factors, variation, disentanglement, inverse, adversarial}

\vskip 0.3in
]



\printAffiliationsAndNotice{}  

\clubpenalty = 10000
\widowpenalty = 10000
\displaywidowpenalty = 10000

\begin{abstract}
Statistical learning relies upon data sampled from a distribution, and we usually do not care what actually generated it in the first place. From the point of view of causal modeling, the structure of each distribution is induced by physical mechanisms that give rise to dependences between observables.
Mechanisms, however, can be meaningful autonomous modules of generative models that make sense beyond a particular entailed data distribution, lending themselves to transfer between problems.
We develop an algorithm to recover a set of independent (inverse) mechanisms from a set of transformed data points. The approach is unsupervised and based on a set of experts that compete for data generated by the mechanisms, driving specialization. We analyze the proposed method in a series of experiments on image data. Each expert learns to map a subset of the transformed data back to a reference distribution. The learned mechanisms generalize to novel domains. We discuss implications for transfer learning and links to recent trends in generative modeling.
\end{abstract}


\section{Introduction}
\label{sec:intro}
Humans are able to recognize objects such as handwritten digits based on distorted inputs.
They can correctly label translated, corrupted, or inverted digits, without having to re-learn them from scratch.
The same applies for new objects, essentially after having seen them once.
Arguably, human intelligence utilizes \emph{mechanisms} (such as translation) that are independent from an input domain and thus generalize across object classes.
These mechanisms are \emph{modular, re-usable and broadly applicable}, and the problem of learning them from data is fundamental for the study of transfer and domain adaptation.

In the field of causality, the concept of independent mechanisms plays a central role both on the conceptual level and, more recently, in applications to inference.
The \emph{independent mechanisms (IM)} assumption  states that the causal generative process of a system's variables is composed of autonomous modules that do not inform or influence each other \citep{SchoelkopfICML2012,PetJanSch17}.

If a joint density is Markovian with respect to a directed graph $\cal G$, we can write it as
\begin{equation}\label{eq:markov}
p({\bf x}) = p(x_1, \ldots, x_d) = \prod_{j = 1}^d p(x_j | {\mbox{pa}_{\cal G}^j}),
\end{equation}
where $\mbox{pa}_{\cal G}^j$ denotes the parents of variable $x_j$ in the graph.

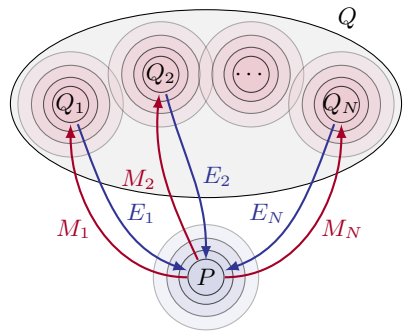
\begin{figure}
\centering
\definecolor{SeaBlue}{RGB}{49, 54, 149}
\definecolor{DarkRed}{RGB}{165, 0, 38}
\def\ecol{SeaBlue}
\def\mcol{DarkRed}
\begin{tikzpicture}[font=\small]
  \foreach \i in {0.35, 0.26, 0.18, 0.12}{
  \pgfmathsetmacro{\opac}{1 - 2*\i}
  \draw[opacity=\opac, fill=\ecol!20] (0,-0.3) circle (2*\i cm);
  }
  \node (P) at (0,-0.3) {$P$};
  
  \node[ellipse, draw, fill=black!5, minimum height=2.5cm, minimum width=5.2cm, label={[anchor=south]25:$Q$}] at (0, 2.01) {};
  \foreach \xi/\yi/\name [count=\counter] in {-1.8/2/{Q_1}, -0.6/2.4/{Q_2}, 0.6/2.4/{\cdots}, 1.8/2/{Q_N}}{
  \begin{scope}[xshift=\xi cm, yshift=\yi cm]
  	\foreach \i in {0.35, 0.26, 0.18, 0.12}{
    \pgfmathsetmacro{\opac}{1 - 2*\i}
    \draw[opacity=\opac, fill=\mcol!20] (0,0) circle (2*\i cm);
    }
    \node (Q\counter) at (0,0) {$\name$};
  \end{scope}
  }
  \draw[->, >=latex, thick, font=\small, \mcol] (P) to[out=180, in=270] node[left, \mcol]{$M_1$} (Q1);
  \draw[->, >=latex, thick, font=\small, \ecol] (Q1) to[out=290, in=160] node[right, \ecol]{$E_1$} (P);
  
  \draw[->, >=latex, thick, font=\small, \mcol] (P) to[out=115, in=265] node[left, \mcol]{$M_2$} (Q2);
  \draw[->, >=latex, thick, font=\small, \ecol] (Q2) to[out=285, in=90] node[right, \ecol]{$E_2$} (P);
  
  
  \draw[->, >=latex, thick, font=\small, \mcol] (P) to[out=0, in=270] node[right, \mcol]{$M_N$} (Q4);
  \draw[->, >=latex, thick, font=\small, \ecol] (Q4) to[out=250, in=20] node[left, \ecol]{$E_N$} (P);
  
\end{tikzpicture}
\caption{An overview of the problem setup.
Given a sample from a canonical distribution $P$, and one from a mixture of transformed distributions $Q_i$ obtained by mechanisms $M_i$ on $P$, we want to learn inverse mechanisms $E_i$ as independent modules.
Modules (or \textit{experts}) compete amongst each other for data points, encouraging specialization.}
\label{fig:cover_figure}
\end{figure}

For a given joint density, there are usually many decompositions of the form~\eqref{eq:markov}, with respect to different graphs.
If $\cal G$ is a \emph{causal} graph, i.e., if its edges denote direct causation \citep{Pearl00}, then the conditional $p(x_j | {\mbox{pa}_{\cal G}}^j)$ can be thought of as physical \emph{mechanism} generating $x_j$ from its parents, and we refer to it as a \emph{causal conditional}.
In this case, we consider the factorization~\eqref{eq:markov} a \emph{ generative} model where the term ``generative'' truly refers to a physical generative process.
As an aside, we note that in the alternative view of causal models as structural equation models, each of the causal conditionals corresponds to a functional mapping and a noise variable \citep{Pearl00}.

By the IM assumption, the causal conditionals are autonomous modules that do not influence or inform each other.
This has multiple consequences.
First, knowledge of one mechanism does not contain information about another one (Appendix~\ref{sec:kc}).
Second, if one mechanism changes (e.g., due to distribution shift), there is no reason that other mechanisms should also change, i.e., they tend to remain \emph{invariant}.
As a special case, it is (in principle) possible to locally \emph{intervene} on one mechanism (for instance, by setting it to a constant) without affecting any of the other modules.
In all these cases, most of~\eqref{eq:markov} will remain unchanged.
However, since the overall density will change, most generic (non-causal) conditionals would change.

The IM assumption can be exploited when performing causal structure inference~\citep{PetJanSch17}.
However, it also has implications for machine learning more broadly.
A model which is expressed in terms of causal conditionals (rather than conditionals with respect to some other factorization) is likely to have components that better transfer or generalize  to other settings~\citep{SchoelkopfICML2012}, and its modules are better suited for building complex models from simpler ones.
Independent mechanisms as sub-components can be trained independently, from multiple domains, and are more likely to be re-usable.
They may also be easier to \emph{interpret} and provide more \emph{insight} since they correspond to physical mechanisms.

Animate intelligence cannot afford to learn new models from scratch for every new task.
Rather, it is likely to rely on robust local components that can flexibly be re-used and re-purposed.
It also requires local mechanisms for adapting and training modules rather than re-training the whole brain every time a new task is learned.
Currently, machine learning excels at optimizing well-defined tasks from large i.i.d.~datasets.
However, if we want to move towards life-long learning and generalization across tasks, then we need to understand how modules can be learnt from data and shared between tasks.

In the present paper, we focus on a class of such modules, and on algorithms to learn them from data.
We describe an architecture using competing experts that automatically specialize on different image transformations.
The resulting model is attractive for lifelong learning, with the possibility of easily adding, removing, retraining, and re-purposing its components independently.
It is unsupervised in the sense that the images are not labeled by the transformations they have undergone.
We only need a sample from a reference distribution and a set of transformed images.
The transformed images are based on another sample, and no pairing or information about the transformations is available.

We test our approach on \emph{MNIST} digits which have undergone various transformations such as contrast inversion, noise addition and translation. Information about the nature and number of such transformations is not known at the beginning of training.
We identify the independent mechanisms linking the reference distribution to a distribution of modified digits, and learn to invert them without supervision.

The inverse mechanisms can be re-purposed as preprocessors, to transform modified digits which are subsequently classified using a standard MNIST classifier.
The trained experts also generalize to \emph{Omniglot} characters, none of which were seen during training.
These are promising results pointing towards a form of robustness that animate intelligence excels at.

\section{Related work}
\label{sec:related}
Our work mainly draws from mixtures of experts, domain adaptation, and causality.

Early works on mixture of experts date back to the early nineties \citep{jacobs1991adaptive,jordan1994hierarchical}, and since then the topic has been subject of extensive research.
Recent work includes that of \citet{shazeer2017outrageously}, successfully training a mixture of 1000 experts using a gating mechanism that selects only a fraction of experts for each example.
\citet{aljundi2017expert} train a network of experts on multiple tasks, with a focus on lifelong learning; autoencoders are trained for each task and used as gating mechanisms.
\citet{NIPS2016_6270} propose Stochastic Multiple Choice Learning, an algorithm which resembles the one we describe in Section \ref{sec:method_compet}, aimed at training mixture of experts to propose a diverse set of outputs.
The main differences are that our model is trained jointly with a \emph{learned} selection system which is valid also at test time, that our trained experts learn independent mechanisms and can be combined (cf.\ Figure~\ref{fig:three_tasks}), and in the way experts are initialized.

Another research direction that is relevant to our work is unsupervised domain adaptation \citep{bousmalis2017unsupervised}.
These methods often use some supervision from labeled data and/or match the two distributions in a learned feature space \citep[e.g.][]{tzeng2017adversarial}.

The novelty of our work lies in the following aspects: (1) we automatically identify and invert a set of independent (inverse) causal mechanisms; (2) we do so using only data from an original distribution and from the mixture of transformed data, without labels; (3) the architecture is modular, can be easily expanded, and its trained modules can be reused; and (4) the method relies on competition of experts.

Ideas from the field of causal inference inspire the present work. Understanding the data generating mechanisms plays a key role in causal inference, and goes beyond the statistical assumptions usually exploited in machine learning.
Causality provides a framework for understanding how a system responds to \emph{interventions}, and causal graphical models as well as structural equation models (SEM) are common ways of describing causal systems~\citep{Pearl00,PetJanSch17}.
The IM assumption discussed in the introduction can be used for identification of causal models~\citep{DanJanMooZscSteZhaSch10,zhang2015distinguishing}, but causality has also proven a useful tool for discussing and understanding machine learning in the non-i.i.d.\ regime.
Recent applications include semi-supervised learning~\citep{SchoelkopfICML2012} and transfer learning~\citep{rojas2015causal}, in which the authors focus only on linear regression models.
We seek to extend applications of causal inference to more complex settings and aim to learn causal mechanisms and ultimately causal SEMs without supervision.

On the conceptual level, our setting is related to recent work on deep learning for disentangling factors of variation \citep{chen2016infogan,higgins2016beta} as well as non-linear ICA \citep{hyvarinen2016unsupervised}.
In our work, causal mechanisms play the role of factors of variation.
The main difference is that we recover mechanisms as independent modules.

\section{Learning causal mechanisms as independent modules}
\label{sec:method_compet}
The aim of this section is twofold.
First, we describe the generative process of our data.
We start with a distribution~$P$ that we will call ``canonical'' and an a priori \emph{unknown} number of independent mechanisms which act on (examples drawn from) $P$.
At training time, a sample from the canonical distribution is available, as well as a dataset obtained by applying the mechanisms to (unseen) examples drawn from~$P$.
Second, we propose an algorithm which recovers and learns to invert the mechanisms in an {unsupervised fashion}.

\subsection{Formal setting}

Consider a canonical distribution~$P$ on~$\mathbb{R}^d$, e.g., the empirical distribution defined by MNIST digits on pixel space.
We further consider~$N$ measurable functions $M_1,\ldots,M_N :\mathbb{R}^d \to \mathbb{R}^d$, called \emph{mechanisms}.
We think of these as independent causal mechanisms in nature, and their number is a priori unknown.
A more formal definition of independence between mechanisms is relegated to Appendix~\ref{sec:kc}.
The mechanisms give rise to~$N$ distributions $Q_1,\ldots,Q_N$ where $Q_j = M_j(P)$.\footnote{Each distribution~$Q_j$ is defined as the pushforward measure of $P$ induced by~$M_j$.}
This setup is illustrated in Figure~\ref{fig:cover_figure}.
In the MNIST example, we consider translations or adding noise as mechanisms,~i.e., the corresponding~$Q$ distributions are translated and noisy MNIST digits.

At training time, we receive a dataset $\mathcal{D}_Q=(x_i)_{i=1}^n$ drawn i.i.d.\ from a mixture of $Q_1,\ldots,Q_N$, and a dataset $\mathcal{D}_P$ sampled independently from the canonical distribution $P$.
Our goal is to identify the underlying mechanisms~$M_1, \ldots, M_N$ and learn approximate inverse mappings which allow us to map the examples from $\mathcal{D}_Q$ back to their counterpart in $P$.

If we were given distinct datasets $\mathcal{D}_{Q_j}$ each drawn from $Q_j$, we could individually learn each mechanism, resulting in independent (approximations of the) mechanisms regardless of the properties of the training procedure.
This is due to the fact that the datasets are drawn from independent mechanisms in the first place, and the separate training procedure cannot generate a dependence between them.
This statement does not require that the procedure is successful, i.e., that the obtained mechanisms approximate the true $M_j$ in some metric.

In contrast, we do not require access to the distinct datasets.
Instead we construct a larger set $\mathcal{D}_Q$ by first taking the union of the sets $D_{Q_j}$, and then applying a random permutation.
This corresponds to a dataset where each element has been generated by one of the (independent) mechanisms, but we do not know by which one.
Clearly, it should be harder to identify and learn independent mechanisms from such a dataset.
We next describe an approach to handle this setting.

\subsection{Competitive learning of independent mechanisms}\label{sec:training}

The training machine is composed of $N'$ parametric functions $E_1,\ldots,E_{N'}$ with distinct trainable parameters $\theta_1,\ldots,\theta_{N'}$.
We refer to these functions as the \emph{experts}.
Note that we do not require $N' = N$, since the real number of mechanisms is unknown a priori.
The goal is to maximize an objective function $c:\mathbb{R}^d\to \mathbb{R}$ with the key property that~$c$ takes high values on the support of the canonical distribution~$P$, and low values outside.
Note that $c$ could be a parametric function, and its parameters could be jointly optimized with the experts during training.
Below, we specify the details of this rather general definition.

\begin{figure}[t]
\centering
\includegraphics[width=0.9\linewidth]{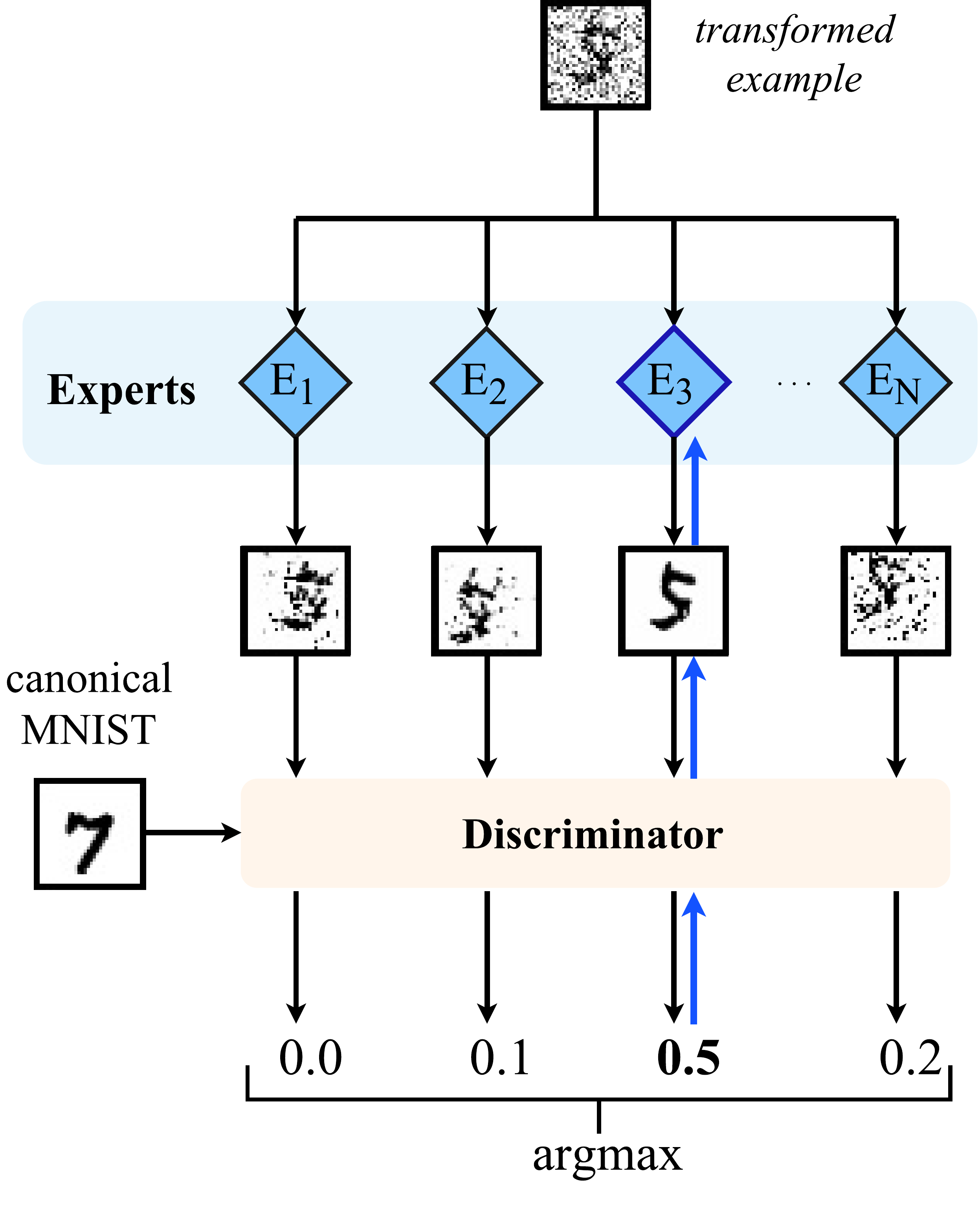}
\caption{We show how a transformed example, here a noisy digit, is processed by a competition of experts.
Only Expert 3 is specializing on denoising, it wins the example and gets trained on it, whereas the others perform translations and are not updated.}
\label{fig:architecture}
\end{figure}

During training, the experts compete for the data points.
Each example $x'$ from $\mathcal{D}_Q$ is fed to all experts independently and in parallel.
Comparing the outputs of all experts $c_j = c(E_j(x'))$, we select the winning expert $E_{j^\ast}$, where $j^\ast =\argmax_j (c_j)$.
Its parameters $\theta_{j^\ast}$ are updated such as to maximize $c(E_{j^\ast}(x'))$, while the other experts remain unchanged.
The motivation behind competitively updating only the winning expert is to enforce specialization; the best performing expert becomes even better at mapping $x'$ back to the corresponding example from the canonical distribution.
We will describe below that alongside with the expert's parameters, we train parameters of $c$ (which in our experiments will be carried in an adversarial fashion).
Figure~\ref{fig:architecture} depicts this procedure.
Overall, our optimization problem reads:
\begin{equation}
\medmuskip=-0.5mu
\thinmuskip=-0.5mu
\thickmuskip=-0.5mu
\nulldelimiterspace=0pt
\scriptspace=1pt
\theta_1^\ast,\ldots,\theta_{N^\prime}^\ast = \argmax_{\theta_1,\ldots,\theta_{N^\prime}} \mathbb{E}_{x^\prime \sim Q}\left(\max_{j\in\{1,\ldots,N^\prime\}}c(E_{\theta_j}(x^\prime))\right).
\end{equation}
The training described above raises a number of questions, which we address next.

\paragraph{1. Selecting the appropriate number of experts.}
Generally, the number of mechanisms $N$ which generated the dataset $\mathcal{D}_Q$ is not available a priori.
Therefore, we require an adaptive procedure to choose the number of experts $N^\prime$.
This is one of the challenges shared with most clustering techniques.
Given the modular behavior of the procedure, experts may be added or removed during or after training, making the framework very flexible.
Assuming however that the number of experts is fixed, the following behaviors could occur.

If $N^\prime > N$ (too many experts):
a) some of the experts do not specialize and do not win any example in the dataset; or
b) some tasks are divided between experts (for instance, each expert can specialize in a mode of the distribution of the same task).
In a), the inactive experts can be removed, and in b) experts sharing the same task can be merged into a wider expert.\footnote{However, note that in order to do this, it is necessary to first acknowledge that the two experts have learned part of the same task, which would require extra information or visual inspection.}

If $N^\prime < N$ (too few experts): a) some of the experts specialize in multiple tasks or b) some of the tasks are not learned by the experts, so that data points from such tasks lead to a poor  score across all experts.
We provide experiments substantiating these claims in \cref{sec:too_many_few}.

\paragraph{2. Convergence criterion.}
Since the problem is  unsupervised, there is no straightforward way of measuring convergence, which raises the question of how to choose a stopping time for the competitive procedure.
As an example, one may act according to one of the following:
\begin{enumerate*}[label=\itshape\alph*\upshape)]
\item fix a maximum number of iterations or
\item stop if each example is assigned to the same experts for a pre-defined number of iterations (i.e., each expert consistently wins the same data points).
\end{enumerate*}

\paragraph{3. Time and space complexity.}
Each example has to be evaluated by all experts in order to assign it to the winning expert.While this results in a computational cost that depends linearly on the number of experts, these evaluations can be done in parallel and therefore the time complexity of a single iteration can be bounded by the complexity to compute the output of a single expert.
Moreover, as each expert will in principle have a smaller architecture than a single large network, the committee of experts will typically be faster to execute.

\paragraph{Concrete protocol for neural networks.}
One possible model class for the experts are deep neural networks.
Training using backpropagation is particularly well suited for the online nature of the training proposed: after an expert wins a data point $x^\prime$, its parameters are updated by backpropagation, while the other experts remain untouched.
Moreover, recent advances in generative modeling give rise to natural choices for the loss function $c$.
For instance, through adversarial training \citep{goodfellow2014generative}, one can use as objective function the output of a discriminator network trained on the canonical sample $\mathcal{D}_P$ and against the outputs of the experts.
In the next section we introduce a formal description of a training procedure based on adversarial training in Algorithm~\ref{alg:alg}, and empirically evaluate its performance.

While in this work we focus on adversarial training, preliminary experiments have shown that similar results can be achieved for example with variational autoencoders (VAE)~\citep{kingma2013auto}.
Given a VAE trained on the canonical distribution $P$, one may define $c(x^\prime)$ as the opposite of the VAE loss.

\section{Experiments}
\label{sec:experiments}
In this set of experiments we test the method presented in Section~\ref{sec:method_compet} on the MNIST dataset transformed with the set of mechanisms described in detail in the Appendix~\ref{sec:details_transf}, i.e., eight directions of translations by 4 pixels (up, down, left, right, and the four diagonals), contrast inversion, addition of noise, for a total of 10 transformations.
We split the training partition of MNIST in half, and transform all and only the examples in the first half; this ensures that there is no matching ground truth in the dataset, and that learning is unsupervised.
As a preprocessing step, the digits are zero-padded so that they have size 32 $\times$ 32 pixels, and the pixel intensities are scaled between 0 and 1.
This is done even before any mechanism is applied.
We use neural networks for both the experts and the selection mechanism, and employ an adversarial training scheme.

Each expert $E_i$ can be seen as a generator from a GAN conditioned on an input image rather than (as usually) a noise vector.
A discriminator $D$ provides gradients for training the experts and acts also as a selection mechanism $c$: only the expert whose output obtains the higher score from $D$ \emph{wins} the example, and is trained on it to maximize the output of $D$.
We describe the exact algorithm used to train the networks in these experiments in Algorithm~\ref{alg:alg}.
The discriminator is trained to maximize the following cross-entropy loss:
\begin{align}
\medmuskip=-0.5mu
\thinmuskip=-0.5mu
\thickmuskip=-0.5mu
\nulldelimiterspace=0pt
\scriptspace=1pt
\begin{split}
\max_{\theta_D}
\biggl(
&\mathbb{E}_{x\sim P}\log(D_{\theta_D}(x))\\
& + \frac{1}{N^\prime}\sum_{j=1}^{N^\prime}\mathbb{E}_{x^\prime \sim Q} \bigr(
\log(1-D_{\theta_D}(E_{\theta_j}(x^\prime)))
\bigr)
\biggr)
\end{split}
\label{eq:crossentropy}
\end{align}

For simplicity, we assume for the rest of this section that the number of experts $N^\prime$  equals the number of true mechanisms $N$.
Results where $N\neq N^\prime$ are relegated to Appendix~\ref{sec:too_many_few}.

\begin{algorithm}[tb]
  \caption{Learning independent mechanisms using competition of experts and adversarial training
  \label{alg:alg}}
  \begin{algorithmic}[1]
    \Require{$X$: data sampled from $P$; $X'$: data sampled from $\mathcal{D}_Q$; $D$ discriminator; $N^\prime$: number of experts; $T$: maximum number of iterations;}
    \Statex{\textbf{(p)} highlights that the steps in the instruction can be executed in parallel}
    \Statex

    \LeftComment{Initialize experts as approximately identity \textbf{(p)}:}
    \Let{$\{E_i$}{TrainAsIdentityOn$(X')\}_{j=1}^{N^\prime}$}

    \For{$t \gets 1 \textrm{ to } T$}

      \LeftComment{Sample minibatches:}
    	\Let{$x, x'$}{Sample($X$), Sample($X'$)}

      \LeftComment{Scores from $D$ for all outputs from the experts \textbf{(p)}:}
  	 	\Let{$\{c_j$}{$D(E_j(x'))\}_{j=1}^{N^\prime}$}

      \LeftComment{Update $D$ \textbf{(p)}:}
      \Let{$\theta_D^{t+1}$}{Adam$\Bigl(\theta_D^{t}, \nabla \log D(x)$}
      \Statex \hspace{3cm}$ + \nabla(1/N^\prime  \sum_{j=1}^{N^\prime} \log (1 - c_j))\Bigr)$

      \LeftComment{Update experts \textbf{(p)}:}
      \Let{$\{\theta_{E_j}^{t+1}$}{Adam$(\theta_{E_j}^{t}, \nabla  \max_{j \in {1,\ldots,N^\prime}} \log (c_j))\}_{j=1}^{N^\prime}$}
    \EndFor
  \end{algorithmic}
\end{algorithm}

\paragraph{Neural nets details.}
Each expert is a CNN with five convolutional layers, 32 filters per layer of size 3 $\times$ 3, ELU \citep{clevert2015fast} as activation function, batch normalization \citep{ioffe2015batch}, and zero padding.
The discriminator is also a CNN, with average pooling every two convolutional layers, growing number of filters, and a fully connected layer with 1024 neurons as last hidden layer.
Both networks are trained using Adam as optimizer \citep{kingma2014adam}, with the default hyper-parameters.\footnote{For the exact experimental parameters and architectures see the Appendix~\ref{sec:details_nets}
or the PyTorch implementation we will release.}

Unless specified otherwise, after a random weight initialization we first train the experts to approximate the identity mapping on our data, by pretraining them on predicting identical input-output pairs randomly selected from the transformed dataset.
This makes the experts start from similar grounds, and we found that this improved the speed and robustness of convergence.
We will refer to this as \emph{approximate identity initialization} for the rest of the paper.

A minibatch of 32 transformed MNIST digits, each transformed by a randomly chosen mechanism, is fed to all experts $E_i$.
The outputs are fed to the discriminator $D$, which computes a score for each of them.
For each example the cross entropy loss in Equation~\eqref{eq:crossentropy} and the resulting gradients are computed only for the output of the highest scoring expert, and they are used to update both the discriminator (when 0 is the target in the cross entropy) and the winning expert (when using 1 as the target).
In order to further support the winning expert, we punish the losing experts by training the discriminator against their outputs as well.
Then, a minibatch of canonical MNIST digit is used in order to update the discriminator with `real' data.
We refer to the above procedure as one \emph{iteration}.

We ran the experiments 10 times with different random seeds for the initializations.
Each experiment is run for 2000 iterations.

\begin{figure}
\centering
\includegraphics[width=\linewidth]{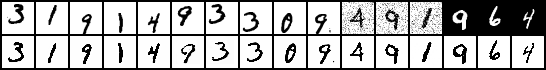}
\caption{The top row contains 16 random inputs to the networks, and the bottom row the corresponding outputs from the highest scoring experts against the discriminator after 1000 iterations.}
\label{fig:outputs}
\end{figure}

\begin{figure*}[t]
\centering
\includegraphics[width=0.9\linewidth]{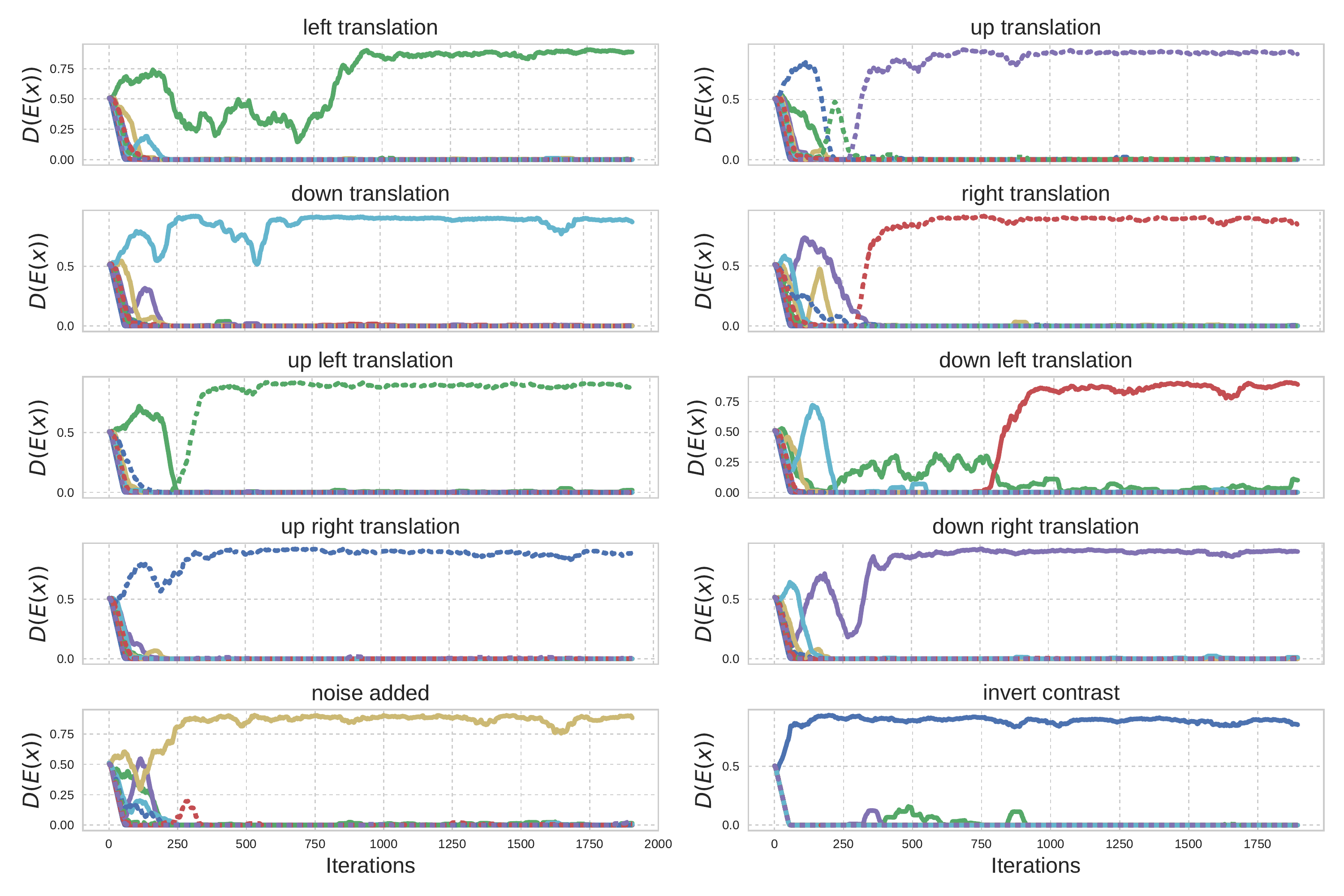}
\caption{Experts' performance, measured by discriminator scores.
Each line color/style represents one expert.
For each of ten different mechanisms (top left to bottom right), the experts are being fed transformed digits.
Each expert learns to specialize on a different mechanism, as shown by the score approaching 1.
Each curve is smoothed with a moving average of 50 iterations.}
\label{fig:plots}
\end{figure*}

\section{Results}
\label{sec:results}
The experts correctly specialized on inverting exactly one mechanism each in $7$ out of the $10$ runs;
in the remaining 3 runs the results were only slightly suboptimal:
one expert specialized on two tasks, one expert did not specialize on any, and the remaining experts still specialized on one task each, thus still covering all the existing tasks.
In Figure~\ref{fig:outputs} we show a randomly selected batch of inputs and corresponding outputs from the model.
Each independent mechanism was inverted by a different expert.

We first discuss our three main findings, and then move on to additional experiments.

\begin{figure}[ptbh]
\centering
\begin{subfigure}{\linewidth}
  \centering
  \includegraphics[width=0.9\linewidth]{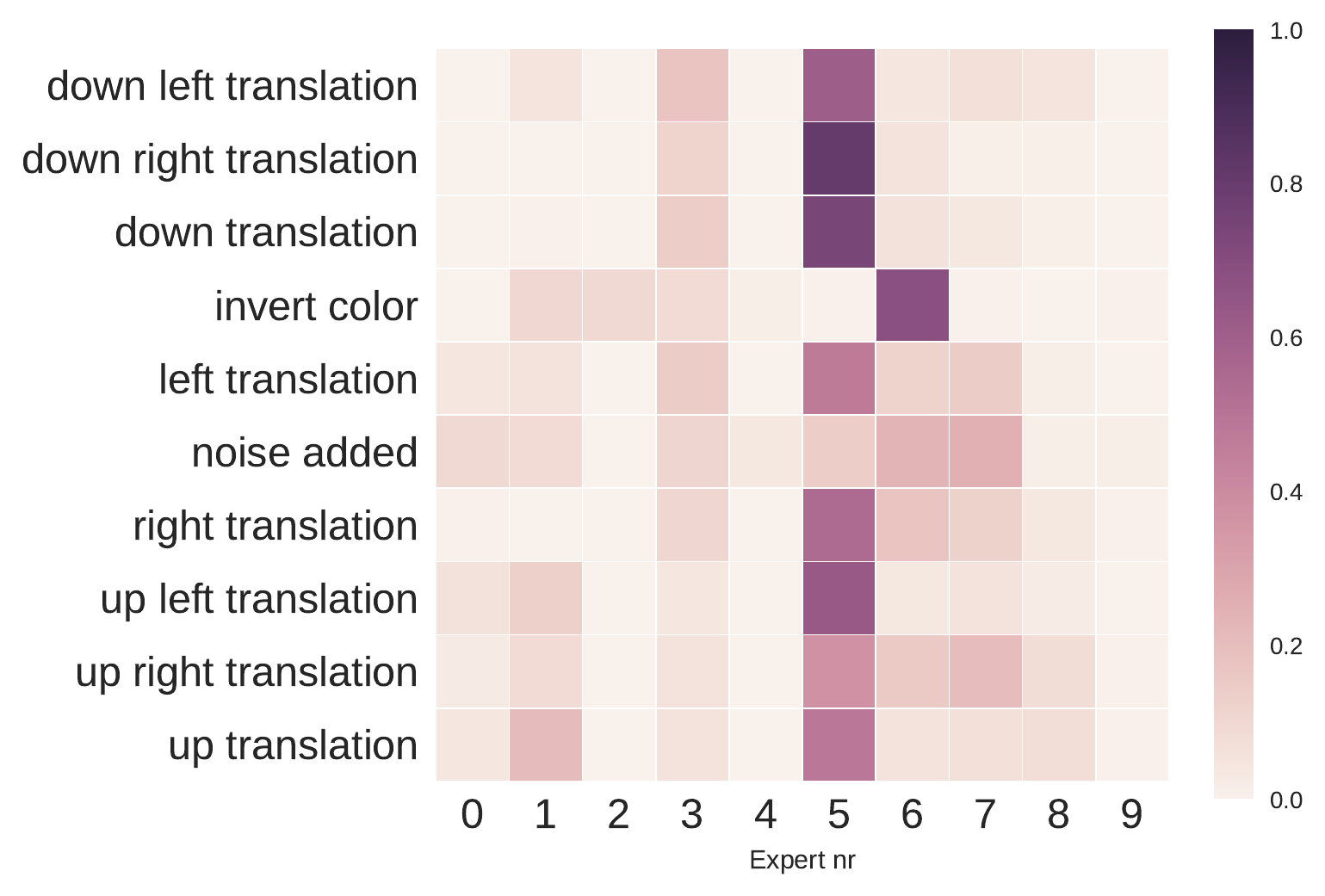}
  \caption{Before training}
\end{subfigure}\\
\begin{subfigure}{\linewidth}
  \centering
  \includegraphics[width=0.9\linewidth]{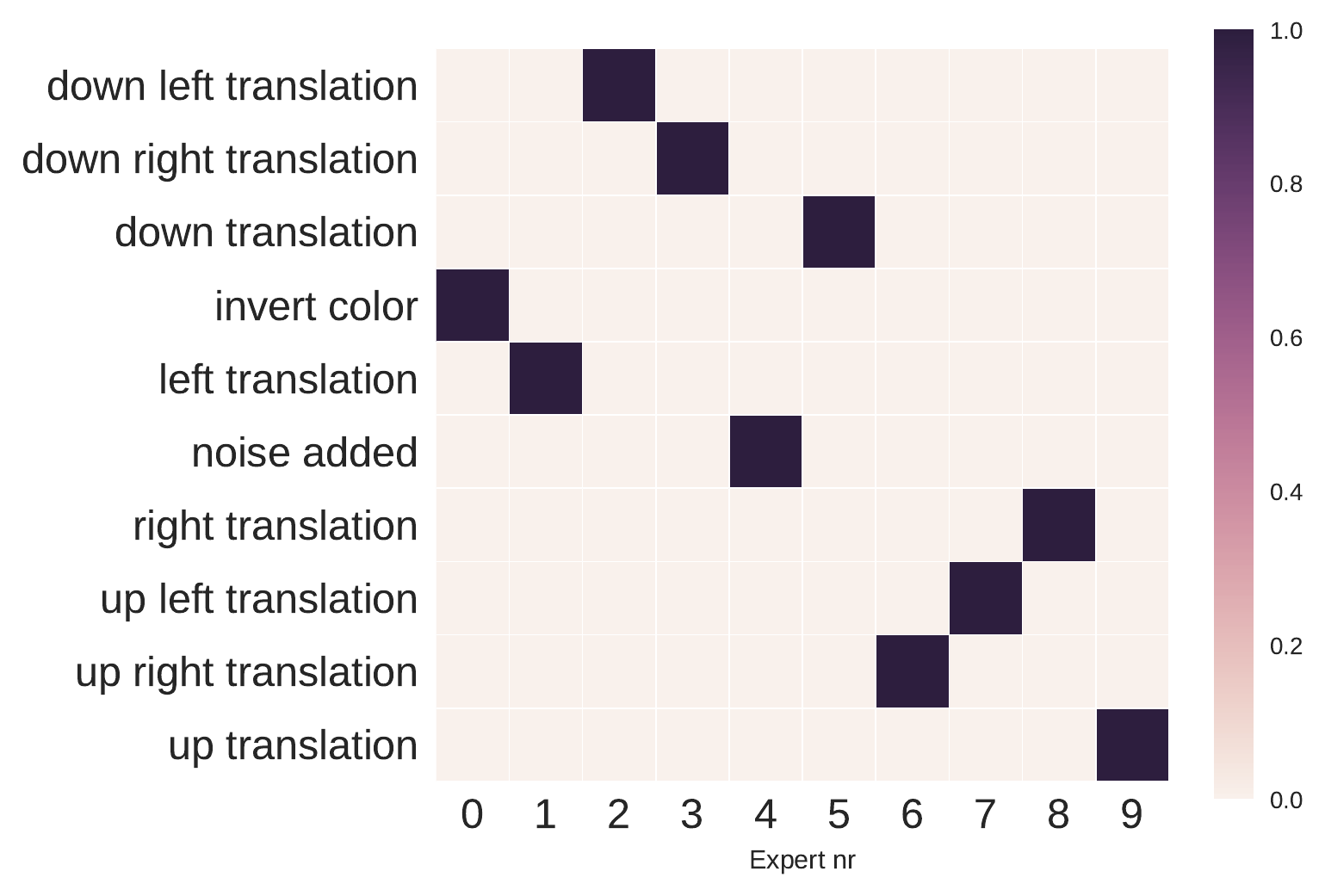}
  \caption{After 1000 iterations}
\end{subfigure}
\caption{The proportion of data won by each expert for each transformation on the digits from the test set.}
\label{fig:matrices}
\end{figure}

\textbf{1. The experts specialize w.r.t.\ $c$.}
In Figure~\ref{fig:plots}, we plot the scores assigned by the discriminator for each expert on each task in a typical successful run.
Each expert is represented with the same color and linestyle across all tasks.
The figure shows that after an initial phase of heavy competition, the experts exhibit the desired behavior and obtain a high score on $D$ on one mechanism each.
Note how the green expert tries to learn two similar tasks until iteration 750 (left and left-down translation), at which point the red expert takes over one of the tasks.
Subsequently, both specialize rapidly.
Figure~\ref{fig:matrices} provides further evidence, by visualizing that the assignments of data points to experts induced by $c$ are meaningful.
We report the proportion of examples from each task assigned to each expert at the beginning and at the end of training:
at first, the assignment of experts to tasks by the discriminator is almost uniform; by the end of the training, each expert wins almost all examples coming from one transformation, and no others.

\textbf{2. The transformed outputs improve a classifier.}
In order to test if the committee of experts can recover a good approximation of the original digits, we test the output of our experts against a pretrained standard MNIST classifier.
For this, we use the test partition of the data.
We compare the accuracy for three inputs:
\begin {enumerate*} [label=\itshape\alph*\upshape)]
\item the test digits transformed by the mechanisms,
\item the transformed digits after being processed by the highest scoring experts (which tries to invert the mechanisms),
\item the original test digits.
The latter can be seen as an upper bound to the accuracy that can be achieved.
\end {enumerate*}

\begin{figure}[ptbh]
\centering
\includegraphics[width=\linewidth]{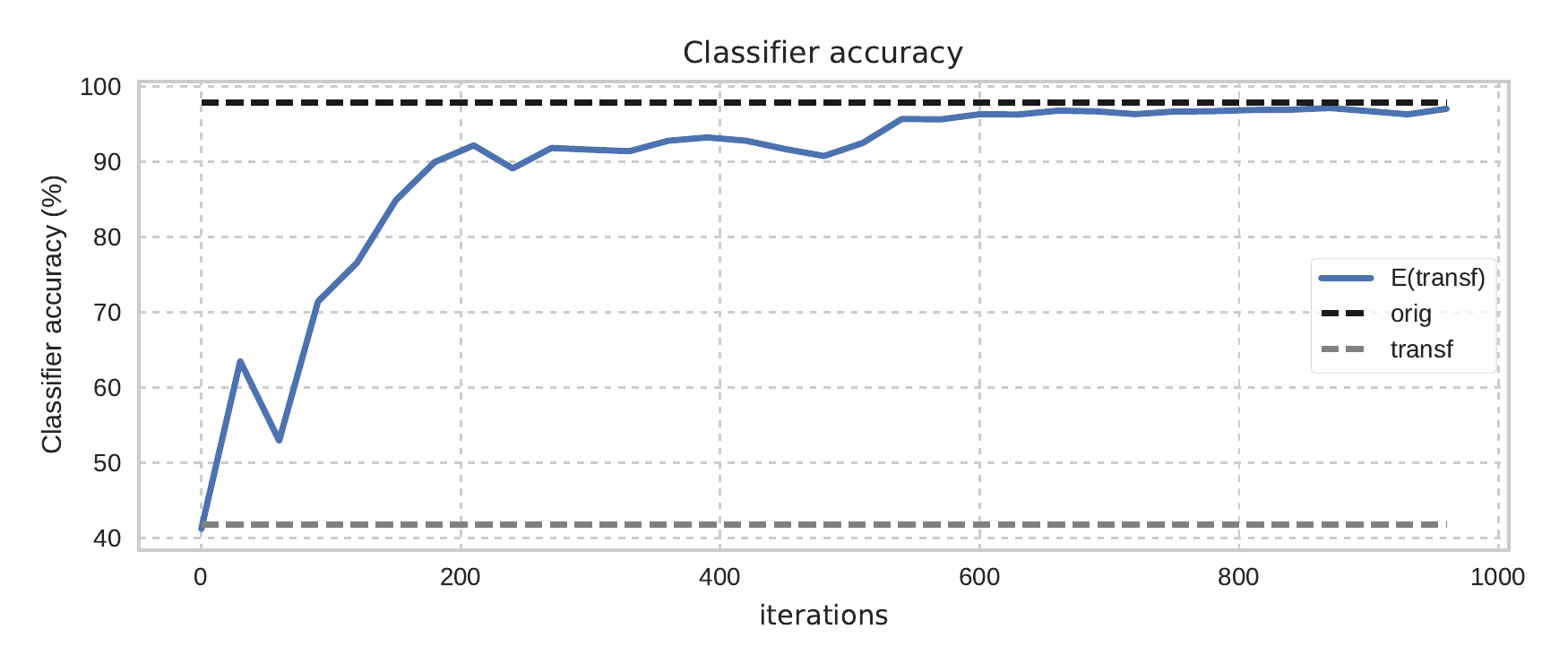}
\caption{Accuracy of a pretrained CNN MNIST classifier on transformed test digits $\mathcal{D}_Q$, on the same digits after going through our model, and on the original digits.
Our system manages to invert the transformations, with the classifier accuracy quickly approaching the optimum.
Note that 600 iterations correspond to having seen about a third of the dataset.}
\label{fig:acc_classif}
\end{figure}

As shown by the two dashed horizontal lines in Figure~\ref{fig:acc_classif}, the transformed test digits achieve a 40\% accuracy when tested directly on the classifier, while the untransformed digits would achieve $\approx 99\%$ accuracy.
The accuracy for the output digits starts at 40\% --- due to the identity initialization of the experts --- but it subsequently quickly approaches the performance on the original digits as it is trained.
Note that after about 600 iterations, i.e., once the networks have seen about one third of the whole dataset \emph{once}, the accuracy has almost reached the upper bound.

\begin{figure}[ptbh]
\centering
\includegraphics[trim={2cm 0 0 0},clip,width=0.9\linewidth]{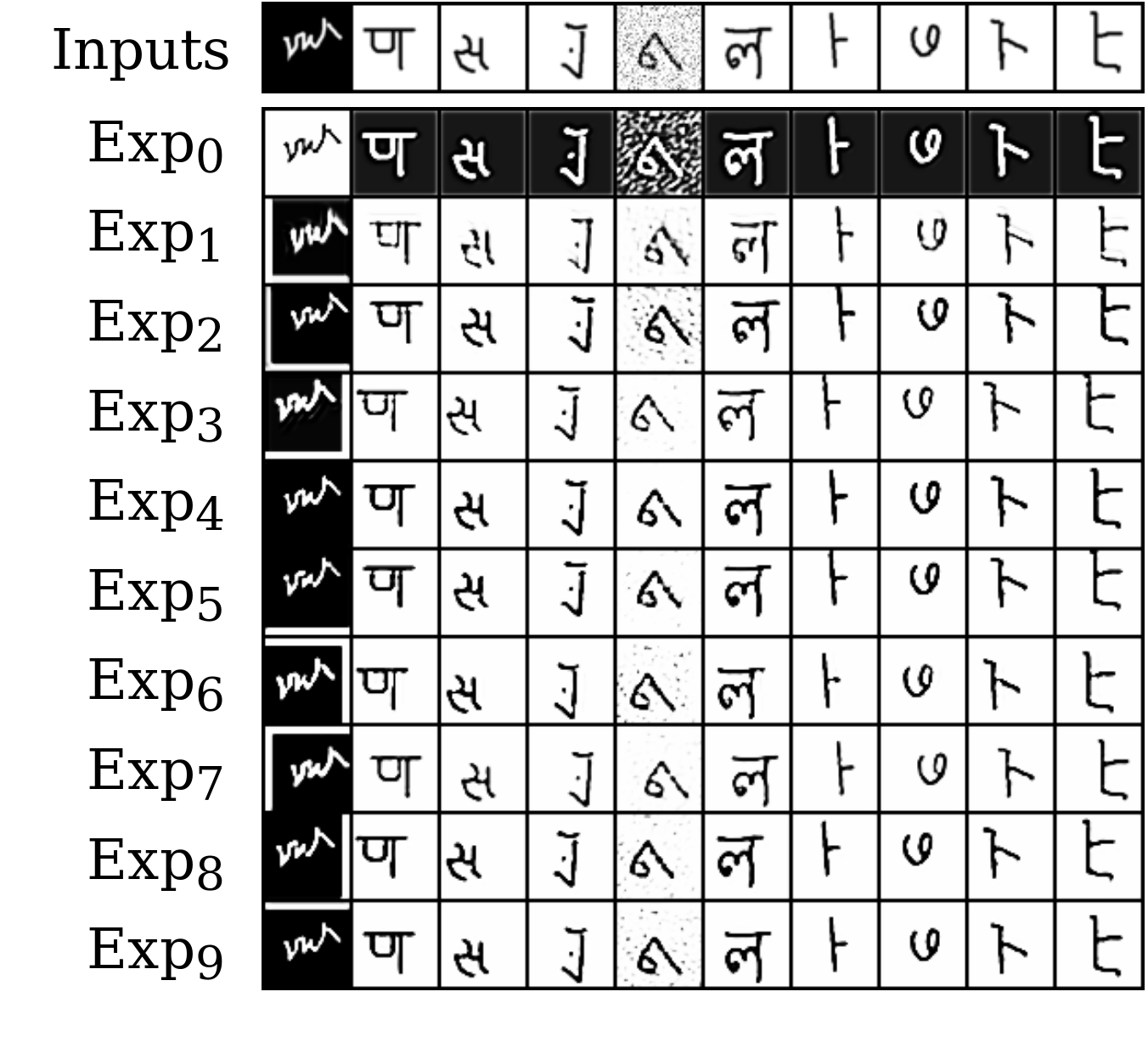}
\caption{Each column shows how each expert transforms the input presented on top.
We arrange the tasks such that the diagonal contains the highest scoring expert for the input given at the top of the column.
The experts have learned the inverse mechanisms, consistently applying them to previously unseen symbols.}
\label{fig:outputs_tasks}
\end{figure}

\textbf{3. The experts learn mechanisms that generalize.}
Given that towards the end of training, each expert~$E_i$ is updated only on data points from~$Q_i$, one could imagine that they will not perform well on data points from other distributions.
In fact this is not the case.
Not only do all experts~$E_i$ generalize to all other transformed distributions~$Q_j$, but also to different datasets all together.
To show this, we use the Omniglot dataset of letters from different alphabets~\citep{lake2015human} and rescale them to $46 \times 46$ pixels (instead of $32 \times 32$ of MNIST, which is not an issue since the experts are fully convolutional).
We transform a random sample with all mechanisms~$M_i$ and test each on all experts~$E_i$, which have only been trained on MNIST.
As shown in Figure~\ref{fig:outputs_tasks}, each network consistently applies the same transformation also on inputs outside of the domain they have specialized on.
They indeed learn a \emph{mechanism} that is independent of the input.

Having made our main points, we continue with a few more observations.

\begin{figure}[ptbh]
\centering
\includegraphics[width=\linewidth]{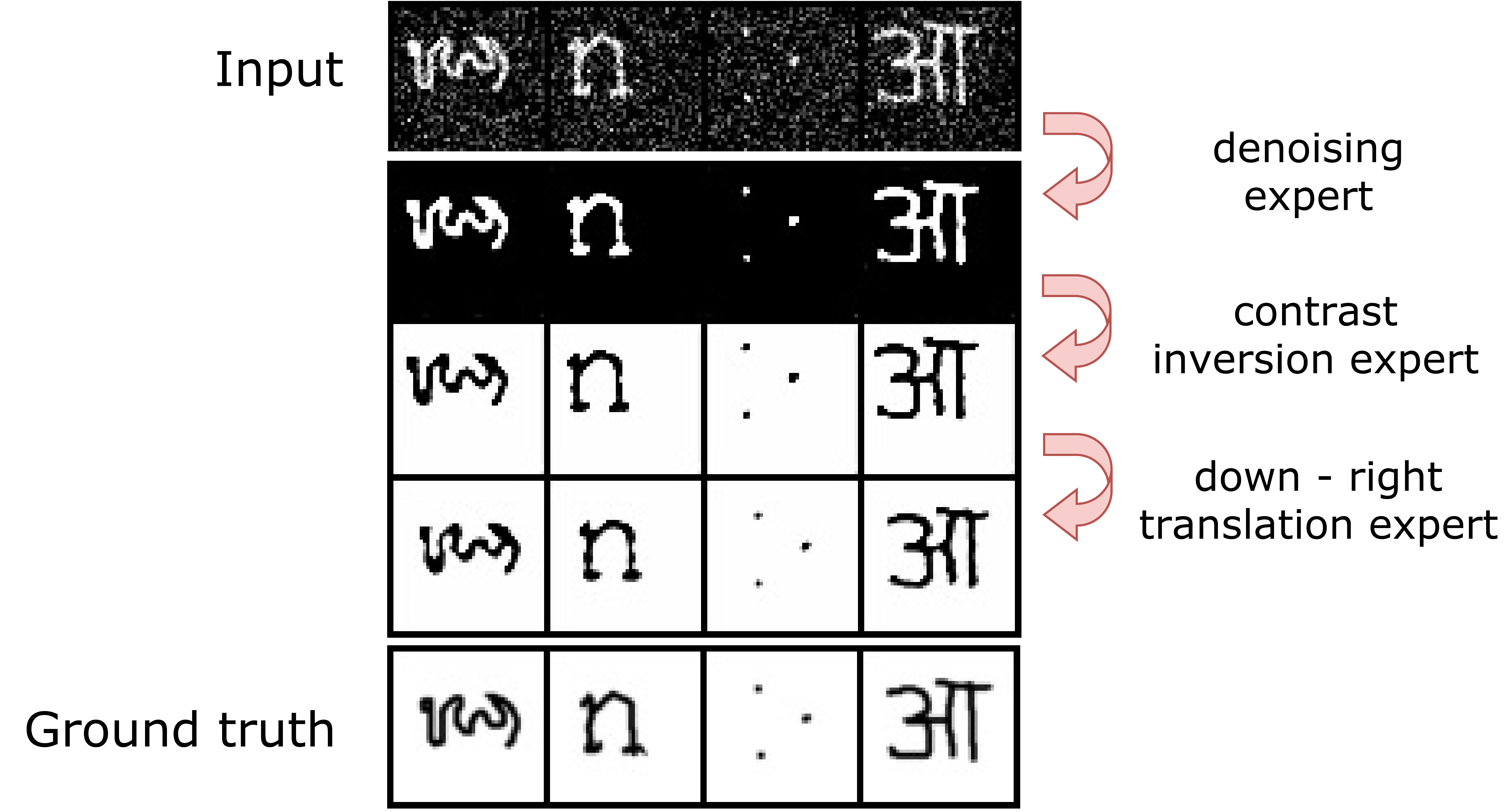}
\caption{First row: input Omniglot letters that were transformed with noise, contrast inversion and translation up left.
Second to fourth row: application of denoising, contrast inverting and right down translating experts.
Last row: ground truth.
Although the experts were not trained on a combination of mechanisms nor on Omniglot letters, they can be used to recover the original digits.}
\label{fig:three_tasks}
\end{figure}

\textbf{The learned inverse mechanisms can be combined.}
We test whether the trained experts could in principle be used to undo several transformations applied at once, even though the training set consisted only of images transformed by a single mechanism.
For simplicity, we assume we know which transformations were used.
In Figure~\ref{fig:three_tasks}, we test on Omniglot letters transformed with three consecutive transformations (noise, up left translation, contrast inversion) by applying the corresponding experts previously trained on MNIST, and correctly recover the original letters.

\paragraph{Effect of the approximate identity initialization.}
For the same experiments but without the approximate identity initialization, several experts fail to specialize.
Out of $10$ new runs with random initialization, only one experiment had arguably good results, with eight experts specializing on one task each, one on two tasks, and the last one on none.
The performance was worse in the remaining runs.
The problem was not that the algorithm takes longer to converge following a random initialization, as with an additional experiment for 10\,000 iterations the results did not improve.
Instead, the random initialization can lead to one expert winning examples from many tasks at the beginning of training, in which case it is hard for the others to catch up.

\paragraph{A simple single-net baseline.}
Training a single network instead of a committee of experts makes the problem more difficult to solve.
Using identical training settings, we trained a single network once with 32, once with 64, and once with 128 filters per layer, and none of them managed to correctly learn more than one inverse mechanism.\footnote{Specifically, the network performs well on the contrast inversion task, and poorly on all others.}
Note that a single network with 128 filters per layer has about twice as many parameters overall as the committee of 10 experts with 32 filters per layer each.
We also tried
\begin {enumerate*} [label=\itshape\alph*\upshape)]
\item random initialization instead of the approximate identity,
\item reducing the learning rate of the discriminator by a factor of 10, and
\item increasing the receptive field by adding two pooling and two upsampling layers, without any improvement.
While we do not exclude that careful hyperparameter tuning may enable a single net to learn multiple mechanisms, it certainly was not straightforward in our experiments.
\end {enumerate*}

\paragraph{Specialization occurs also with higher capacity experts.}
While in principle with infinite capacity and data, a single expert could solve all tasks simultaneously, in practice limited resources and the proposed training procedure favor specialization in independent modules.
Increasing the size of the experts from 32 filters per layer to 64 or 128 filters,\footnote{Equivalent to an increase of parameters from $\sim$27K to $\sim$110K or $\sim$440K parameters respectively.} or enlarging the overall receptive field by using two pooling and two upsampling layers, still resulted in good specialization of the experts, with no more than two experts specializing on two tasks at once.

\paragraph{Fewer examples from the canonical distribution.}
In some applications, we might only have a small sample from the original distribution.
Interestingly, if we reduce the number of examples from the original distribution from 30\,000 down to 64, we find that all experts still specialize and recover good approximations of the inverse mechanisms, using the exact same training protocol.
Although the output digits turn out less clean and sharp, we still achieve 96\% accuracy on the pretrained MNIST classifier.

\section{Conclusions}
\label{sec:conclusion}
We have developed a method to identify and learn a set of independent causal mechanisms.
Here these are \emph{inverse} mechanisms, but an extension to forward mechanisms appears feasible and worthwhile.
We reported promising results in experiments using image transformations; future work could study more complex settings and diverse domains.
The method does not explicitly minimize a measure of dependence of mechanisms, but works if the data generating process contains independent mechanisms in the first place: As the different tasks (mechanisms) do not contain information about each other, improving on one of them does not improve performance on another, which is exactly what encourages specialization.

A natural extension of our work is to consider independent mechanisms that \emph{simultaneously} affect the data (e.g., lighting and position in a portrait), and to allow multiple passes through our committee of experts to identify local mechanisms (akin to Lie derivatives) from more complex datasets --- for instance, using recurrent neural networks that allow the application of multiple mechanisms by iteration.
With many experts, the computational cost (or parallel processing) might become unnecessarily high.
This could be mitigated by hybrid approaches incorporating gated mixture of experts or a hierarchical selection of competing experts.

We believe our work constitutes a promising connection between causal modeling and deep learning.
As discussed in the introduction, causality has a lot to offer for crucial machine learning problems such as transfer or compositional modeling.
Our systems sheds light on these issues.
Independent modules as sub-components could be learned using multiple domains or tasks, added subsequently, and transferred to other problems.
This may constitute a step towards causally motivated life-long learning.

\bibliography{references}
\bibliographystyle{icml2018}

\clearpage
\appendix
\section{Additional results.}

\subsection{Too many or too few experts.}
\label{sec:too_many_few}

\begin{figure}[b]
\centering
\begin{subfigure}{0.73\linewidth}
  \centering
  \includegraphics[width=\linewidth]{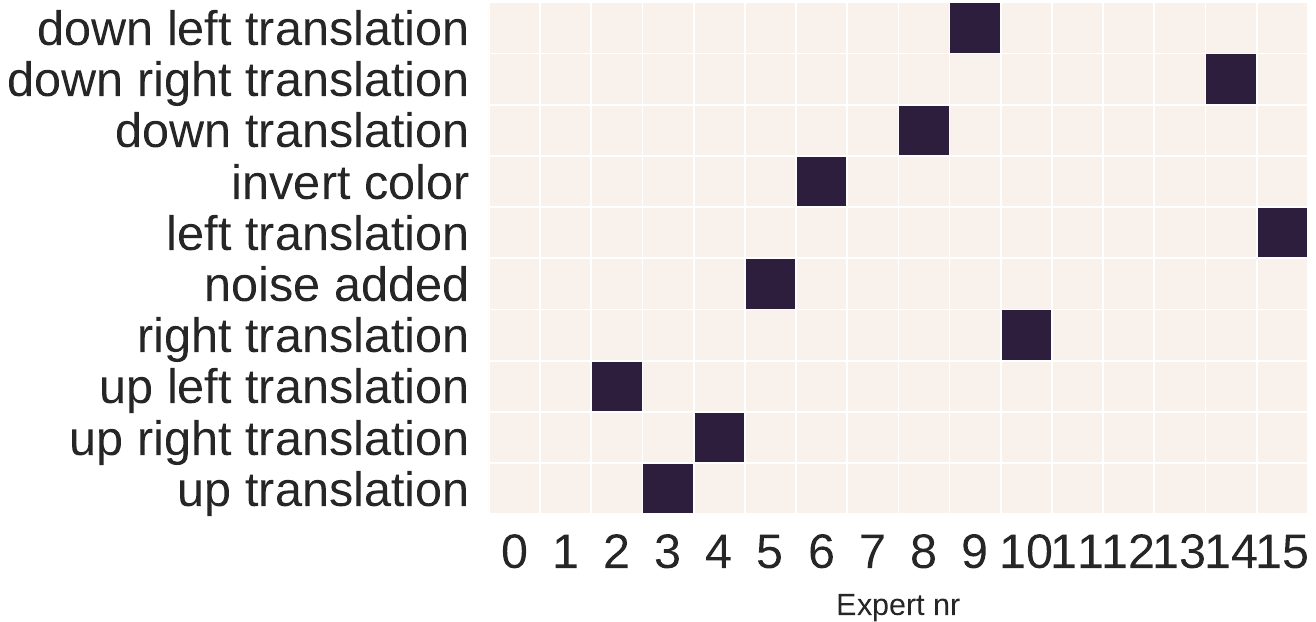}
\end{subfigure}
\begin{subfigure}{0.23\linewidth}
	\centering
  \includegraphics[width=\linewidth]{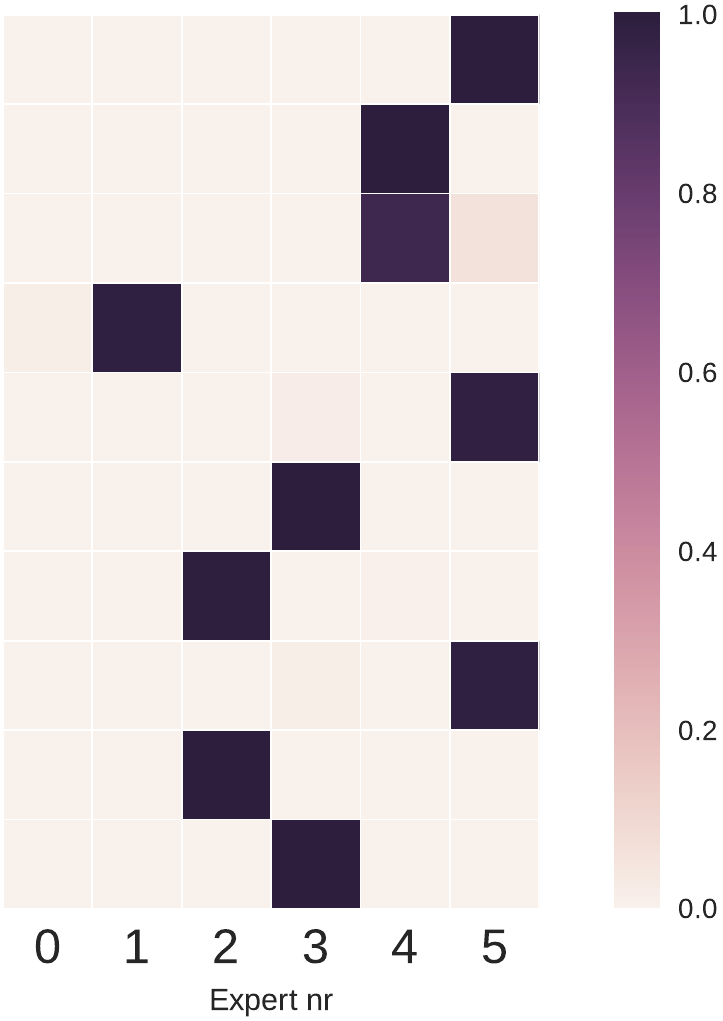}
\end{subfigure}
\caption{The proportion of data won by each expert for each transformation on the digits from the test set, for the case of 10 mechanisms and more experts (16 on left) or too few (6 on the right).
Note how on the left experts 0, 1, 7, 11, 12, 13, do not win any data points, and can therefore be discarded.}
\label{fig:matrices_16_6}
\end{figure}

\paragraph{Too many experts.}
When there are too many experts, for most tasks only one wins all the examples, as shown in Figure~\ref{fig:matrices_16_6} where the model has 16 experts for 10 tasks.
In this case the remaining experts do not specialize at all and therefore can be removed from the architecture.
Had several experts specialized on the same task, they could be combined after determining that they perform the same task.
Since the accuracy on the transformed data tested on the pretrained classifier reaches again the upperbound of the untransformed data, and since the progress is very similar to that illustrated in Figure~\ref{fig:acc_classif}, we omit this plot.

\paragraph{Too few experts.}
For a committee of 6 experts, the networks do not reconstruct properly most of the digits, which is reflected by an overall low objective function value on the data.
Also, the accuracy achieved by the pretrained MNIST classifier does not exceed 72\%.
A few experts are inevitably assigned to multiple tasks, and by looking at Figure~\ref{fig:matrices_16_6} it is interesting to see that the clustering result is still meaningful (e.g., expert 5 is assigned to left, down-left, and up-left translation).

\section{Details of neural networks}
\label{sec:details_nets}

In Table~\ref{tab:D} we report the configuration of the neural networks used in these experiments.

For the approximate identity initialization we train each network for a maximum of 500 iterations, or until the mean squared error of the reconstructed images is below 0.002.

\section{Transformations}
\label{sec:details_transf}

In our experiments we use the following transformations
\begin{itemize}
\item Translations: the image is shifted by 4 pixels in one of the eight directions up, down, left, right and the four diagonals.
\item Contrast (or color) inversion: the value of each pixel --- originally in the range $[0, 1]$ --- is recomputed as $1 -$ the original value.
\item Noise addition: random Gaussian noise with zero mean and variance 0.25 is added to the original image, which is then clamped again to the $[0, 1]$ interval.
\end{itemize}

\begin{table}[tbh]
  \caption{Architectures of the neural networks used in the experiment section.
  BN stands for Batch normalization, FC for fully connected.
  All convolutions are preceded by a 1 pixel zero padding.}
  \centering
  \begin{minipage}{0.47\linewidth}
    \begin{tabular}{c}
      \multicolumn{1}{c}{\textbf{Expert}}\\
      \toprule
      \textbf{Layers} \\ \midrule
      \rowcolor{LightGray}
      $3 \times 3, 32$, BN, ELU\\
      \rowcolor{LightGray}
      $3 \times 3, 32$, BN, ELU\\
      \rowcolor{LightGray}
      $3 \times 3, 32$, BN, ELU\\
      \rowcolor{LightGray}
      $3 \times 3, 32$, BN, ELU\\
      $3 \times 3, 1$, sigmoid\\
      \bottomrule
    \end{tabular}
    \label{tab:E}
  \end{minipage}
  \hfill
  \begin{minipage}{0.47\linewidth}
    \begin{tabular}{c}
      \multicolumn{1}{c}{\textbf{Discriminator}}\\
      \toprule
      \textbf{Layers} \\ \midrule
      $3 \times 3, 16$, ELU\\
      \rowcolor{LightGray}
      $3 \times 3, 16$, ELU\\
      \rowcolor{LightGray}
      $3 \times 3, 16$, ELU\\
      \rowcolor{Gray}
      $2 \times 2$, avg pooling\\
      \rowcolor{LightGray}
      $3 \times 3, 32$, ELU\\
      \rowcolor{LightGray}
      $3 \times 3, 32$, ELU\\
      \rowcolor{Gray}
      $2 \times 2$, avg pooling\\
      \rowcolor{LightGray}
      $3 \times 3, 64$, ELU\\
      \rowcolor{LightGray}
      $3 \times 3, 64$, ELU\\
      \rowcolor{Gray}
      $2 \times 2$, avg pooling\\
      $1024$, FC, ELU \\
      $1$, FC, sigmoid \\
      \bottomrule
    \end{tabular}
    \label{tab:D}
  \end{minipage}
\end{table}

\section{Notes on the Formalization of Independence of Mechanisms}
\label{sec:kc}

In this section we briefly discuss the notion of independence of mechanisms as in \citet{JanSch10}, where the independence principle is formalized in terms of algorithmic complexity (also known as Kolmogorov complexity).
We summarize the main points needed in the present context.
We parametrize each \emph{mechanism} by a bit string $x$.
The Kolmogorov complexity $K(x)$ of $x$ is the length of the shortest program generating $x$ on an a priori chosen universal Turing machine.
The \textbf{algorithmic mutual information} can be defined as $I(x:y):= K(x)+K(y)-K(x,y)$, and it can be shown to equal
\begin{equation}\label{eq:MI}
I(x:y)=K(y) - K(y|x^*),
\end{equation}
where for technical reasons we need to work with $x^*$, the shortest description of $x$ (which is in general uncomputable).
Here, the conditional Kolmogorov complexity $K(y|x)$ is defined as the length of the shortest program that generates $y$ from $x$.
The algorithmic mutual information measures the algorithmic information two objects have in common.
We define two mechanisms to be \textbf{(algorithmically) independent} whenever the length of the shortest description of the two bit strings together is not shorter than the sum of the shortest individual descriptions (note it cannot be longer), i.e., if their algorithmic mutual information vanishes.\footnote{All statements are valid up to additive constants, linked to the choice of a Turing machine which produces the object (bit string) when given its compression as an input. For details, see~\citet{JanSch10}.}
In view of~\eqref{eq:MI}, this means that
\begin{equation}\label{eq:IND}
K(y) = K(y|x^*).
\end{equation}

We will say that two mechanisms $x$ and $y$ are independent whenever the complexity of the conditional mechanism $y|x$ is comparable to the complexity of the unconditional one $y$.
If, in contrast, the two mechanisms were closely related, then we would expect that we can mimic one of the mechanisms by applying the other one followed by a low complexity conditional mechanism.

\end{document}